\documentclass[12pt,margin=1in]{article}
\pdfoutput=1 

\usepackage[round]{natbib}

\usepackage{rotating}
\usepackage{graphicx}

\usepackage{amsthm,amsmath} 
\usepackage{amssymb}
\usepackage{subcaption}
\begin{document}
\thispagestyle{empty}
\baselineskip=28pt

\begin{center}
{\LARGE{\bf BET: Bayesian Ensemble Trees\\ for Clustering and Prediction in Heterogeneous Data }}
\end{center}

\baselineskip=12pt

\vskip 2mm
\begin{center}
Leo L. Duan
\footnote{\baselineskip=10pt Department of Mathematical Sciences, University of Cincinnati},
John P. Clancy
\footnote{\baselineskip=10pt Division of Pulmonary Medicine, Cincinnati Children's Hospital Medical Center}
and
Rhonda D. Szczesniak
\footnote{\baselineskip=10pt Division of Biostatistics and Epidemiology, Cincinnati Children's Hospital Medical Center}
\footnote{\baselineskip=10pt Corresponding author. Address: 3333 Burnet Ave, MLC 5041, Cincinnati, OH 45229. Phone:(513)803-0563, email: rhonda.szczesniak@cchmc.org}
\end{center}
%
%
%
%%%%%%%%%%%%%%%%%%%%%%%%%%%%%%%%%%%%%%%%%%%%%%%%%%%%%%%%%%%%%%%%%%%%%%%%
\begin{center}
{\Large{\bf Summary}}
\end{center}
\baselineskip=12pt

We propose a novel ``tree-averaging'' model that utilizes the ensemble of classification and regression trees (CART). Each constituent tree is estimated with a subset of similar data. We treat this grouping of subsets as Bayesian ensemble trees (BET) and model them as an infinite mixture Dirichlet process. We show that BET adapts to data heterogeneity and accurately estimates each component. Compared with the bootstrap-aggregating approach, BET shows improved prediction performance with fewer trees. We develop an efficient estimating procedure with improved sampling strategies in both CART and mixture models. We demonstrate these advantages of BET with simulations, classification of breast cancer and regression of lung function measurement of cystic fibrosis patients.

%%%%%%%%%%%%%%%%%%%%%%%%%%%%%%%%%%%%%%%%%%%%%%%%%%%%%%%%%%%%%%%%%%%%%%%%
%
%
%

\baselineskip=12pt
\par\vfill\noindent
\par\medskip\noindent

\clearpage\pagebreak\newpage
\pagenumbering{arabic}

\section{Introduction}
Classification and regression trees (CART) \citep*{cart84} is a nonparametric learning approach that provides fast partitioning of data through the binary split tree and an intuitive interpretation for the relation between the covariates and outcome. Aside from simple model assumptions, CART is not affected by potential collinearity or singularity of covariates. From a statistical perspective, CART models the data entries as conditionally independent given the partition, which not only retains the likelihood simplicity but also preserves the nested structure.

Since the introduction of CART, many approaches have been derived with better model parsimony and prediction. The Random Forests model \citep*{breiman2001random} generates bootstrap estimates of trees and utilizes the bootstrap-aggregating (``bagging") estimator for prediction. Boosting \citep*{friedman2001greedy,friedman2002stochastic} creates a generalized additive model of trees and then uses the sum of trees for inference. Bayesian CART \citep*{denison1998bayesian, chipman1998bayesian} assigns a prior distribution to the tree and uses Bayesian model averaging to achieve better estimates. Bayesian additive regression trees (BART, \cite*{chipman2010bart}) combine the advantages of the prior distribution and sum-of-trees structure to gain further improvement in prediction.

Regardless of the differences in the aforementioned models, they share one principle: multiple trees create more diverse fitting than a single tree; therefore, the combined information accommodates more sources of variability from the data. Our design follows this principle.

We create a new ensemble approach called the Bayesian Ensemble Trees (BET) model, which utilizes the information available in the subsamples of data. Similar to Random Forests, we hope to use the average of the trees, of which each tree achieves an optimal fit without any restraints. Nonetheless, we determine the subsamples through clustering rather than bootstrapping. This setting automates the control of the number of trees and also adapts the trees to possible heterogeneity in the data.

In the following sections, we first introduce the model notation and its sampling algorithm. We illustrate the clustering performance through three different simulation settings. We demonstrate the new tree sampler using homogeneous example data from a breast cancer study. Next we benchmark BET against other tree-based methods using heterogeneous data on lung function collected on cystic fibrosis patients. Lastly, we discuss BET results and possible extensions.

\section{Preliminary Notation}
We denote the $i$th record of the outcome as $Y_i$, which can be either categorical or continuous. Each $Y_i$ has a corresponding covariate vector ${\bf X}_i$.

In the standard CART model, we generate a binary decision tree $\bf T$ that uses only the values of $\bf X_i$ to assign the $i$th record to a certain region. In each region, elements of $Y_i$ are identically and independently distributed with a set of parameters ${\boldsymbol \theta}$. Our goals are to find the optimal tree $\bf T$, estimate ${\boldsymbol \theta}$ and make inference about an unknown $Y_s$ given values of ${\bf X}_s$, where $s$ indexes the observation to predict.

We further assume that $\{Y_i, \bf X_i\}$ is from one of (countably infinitely) many trees $\{{\bf T}_j\}_j$. Its true origin is only known up to a probability $w_j$ from the $j$th tree. Therefore, we need to estimate both ${\bf T}_j$ and $w_j$ for each $j$. Since it is impossible to estimate over all $j$'s, we only calculate those $j$'s with non-negligible $w_j$, as explained later.

 %Each tree $\bf T$ sets up the splitting criteria in $X$ and partitions the parameter space for $\theta$. Then $i$th subject is passed down the $j$th tree $T_j$ to the partition $R_{(j)k}$,in a deterministic manner based on ${\bf X}_i$. The value of $\theta$ in $R_{(j)k}$ is then used as the predictor for $Y_i$. From the probabilistic point of view, likelihood is formulated with $Y_i$ and $\theta$ in $R_{(j)k}$ and independence is assumed for different $i$'s. Based on the type of data, the likelihood takes the common form of either continuous (such as normal with $\theta=(\mu,\sigma^2)$) or discrete distribution (such as Bernoulli with $\theta=p$). We use $G$ to denote this distribution.

\section{Bayesian Ensemble Tree (BET) Model}
We now formally define the proposed model. We use $[.]$ to denote the probability density. Let $[Y_i|{\bf X}_i,{\bf T}_j]$ denote the probability of $Y_i$ conditional on its origination is from the $j$th tree. The mixture likelihood can be expressed as:

\begin{equation}
[Y_i|{\bf X}_i,\{{\bf T}_j\}_j]= \sum_j^{\infty} w_j[Y_i|{\bf X}_i,{\bf T}_j]
\label{eqn:bet_model}
\end{equation}
where the mixture weight vector has an infinite-dimension Dirichlet distribution with precision parameter $\alpha$: ${\bf W}=\{w_j\}_j\sim Dir_\infty(\alpha)$. The likelihood above corresponds to $Y_i \stackrel{iid}{\sim} DP(\alpha, G))$, where $DP$ stands for the Dirichlet process and the base distribution $G$ is simply $[Y_i|{\bf T}_j]$.

\subsection{Hierarchical Prior for $\bf T$}

We first define the nodes as the building units of a tree. We adopt the notation introduced by \cite*{wu2007bayesian} and use the following method to assign indices to the nodes: for any node $k$, two child nodes are indexed as left $(2k+1)$ and right $(2k+2)$; the root node has index $0$. The parent of any node $k>0$ is simply $\lfloor\frac{k-1}{2}\rfloor$, where $\lfloor.\rfloor$ denotes the integer part of a non-negative value. The depth of a node $i$ is $\lfloor\log_2 (k+1)\rfloor$. 

 Each node can have either zero (not split) or two (split) child nodes. Conditional on the parent node being split, we use $s_k=1$ to denote one node being split (or interior),  $s_k=0$ otherwise (or leaf). Therefore, the frame of a tree with at least one split is:
 $$[s_0=1]\prod_{k\in I_k}[s_{2k+1}|s_k=1][s_{2k+2}|s_k=1]$$
 where  $I_k=\{k: s_k=1\}$ denotes the set of interior nodes.
 
Each node has splitting thresholds that correspond to the $m$ covariates in ${\bf X}$. Let the $m-$dimensional vector ${\bf t}_k$ denote these splitting thresholds. Also, it has a random draw variable $c_k$ from $\{1,...,m\}$. We assume $s_k, {\bf t}_k, c_k$ are independent.

For the $c_k$th element $X_{(c_k)}$, if $X_{(c_k)}<t_{(c_k)}$, then observation $i$ is distributed to its left child; otherwise it is distributed to the right child. For every $i$, this distributing process iterates from the root node and ends in a leaf node. We use $\theta_k$ to denote the distribution parameters in the leaf nodes. For each node and a complete tree, their prior densities are:
\begin{equation}
\begin{aligned}
&[T_k]=[s_k][c_k]^{s_k}[{\bf t}_k]^{s_k}[\theta_k]^{1-s_k}\\
&[{\bf T}]=[T_0]\prod_{k\in I_k} [T_{2k+1}][T_{2k+2}]
\end{aligned}
\end{equation}

For $s_k, {\bf t}_k, c_k$, we specify the prior distributions as follows:
\begin{equation}
\begin{aligned}
& s_k\sim {B} ( exp(-\lfloor\log_2 (k+1)\rfloor/\delta))\\
& c_k \sim {MN}_m(\boldsymbol \xi) \text{, where }  \boldsymbol \xi\sim Dir (\boldsymbol 1_m)\\
& [{\bf t}_k]\propto 1\\
\end{aligned}
\end{equation}
where $B$ is Bernoulli distribution, $MN_m$ is $m$-dimensional multinomial distribution. The hyper-parameter $\delta$ is the tuning parameter for which smaller $\delta$ results in smaller tree.

In each partition, objective priors are used for $\boldsymbol{\theta}$. If $\bf Y$ is continuous, then $[{\boldsymbol\theta}_k]=[\mu,\sigma^2]\propto 1/\sigma^2$; if $\bf Y$ is discrete, then ${\boldsymbol\theta}_k={\bf p}\sim Dir(0.5\cdot \bf{1})$. Note that $Dir$ reduces to a $Beta$ distribution when $\bf Y$ is a binary outcome. To guarantee the posterior propriety of $\boldsymbol{\theta}$, we further require that each partition should have at least $q$ observations and $q>1$.

The posterior distribution of $\boldsymbol{\xi}$ reveals the proportion of instances that a certain variable is used in constructing a tree. One variable can be used more times than another, therefore resulting in a larger proportion in $\boldsymbol{\xi}$. Therefore, $\boldsymbol{\xi}$ can be utilized in variable selection and we name it as variable ranking probability.

\subsection{Stick-Breaking Prior for $\bf W$}

The changing dimension of the Dirichlet process creates difficulties in Bayesian sampling. Pioneering studies include exploring infinite state space with the reversible-jump Markov chain Monte Carlo \citep*{green2001modelling} and with an auxiliary variable for possible new states \citep*{neal2000markov}. At the same time, an equivalent construction named the stick-breaking process \citep*{ishwaran01} gained popularity for decreased computational burden. The stick-breaking process decomposes the Dirichlet process into an infinite series of $Beta$ distributions:

\begin{equation}
\begin{aligned}
& w_1=v_1\\
& w_j=v_j\prod_{k<j}(1-v_k) \text{ for }j>1
\end{aligned}
\end{equation}
where each $v_j\stackrel{iid}{\sim} Beta(1,\alpha)$. This construction provides a straightforward illustration on the effects of adding/deleting a new cluster to/from the existing clusters.

Another difficulty in sampling is that $j$ is infinite. \cite*{ishwaran01}  proved that the $\max(j)$ can be truncated to $150$ for a sample size of $n=10^5$, and the results are indistinguishable from those obtained using larger numbers. Later, \cite*{kalli2011slice} introduced the slice sampler, which avoids the approximate truncation. Briefly, the slice sampler adds a latent variable $u_i \sim U(0,1)$ for each observation. The probability in (\ref{eqn:bet_model}) becomes:

\begin{equation}
[Y_i|{\bf X}_i,\{{\bf T}_j\}_j]= \sum_j^{\infty} \boldsymbol{1}(u_i<w_j )[Y_i|{\bf X}_i,{\bf T}_j]
\end{equation}
due to $\int_0^1\boldsymbol{1}(u_i<w_j )d{u_i}=w_j$. The Monte Carlo sampling of $u_i$ leads to  omitting $w_j$'s that are too small. We found that the slice sampler usually leads to a smaller effective $\max{j}<10$ for $n=10^5$, hence more rapid convergence than a simple truncation.

\section{Blocked Gibbs Sampling}

We now explain the sampling algorithm for the BET model. Let $Z_i=j$ denote the latent assignment of the $i$th observation to the $j$th tree. Then the sampling scheme for the BET model involves iteration over two steps: tree growing and clustering.

\subsection{Tree Growing: Updating $[\bf T|W,Z, Y]$}
Each tree with allocated data is grown in this step.  We sample in the order of $[\bf s,c,t]$ and then $[\bf {\boldsymbol{\theta}}|s,c,t]$. As stated by \cite*{chipman1998bayesian}, using $[\bf Y|s,c,t]$ marginalized over $\boldsymbol \theta$ facilitates rapid change of the tree structure. After the tree is updated, the conditional sampling of $\boldsymbol{\theta}$ provides convenience in the next clustering step, where we compute $[Y_i|{\bf T}_j]$ for different $j$.

During the updating of $[\bf s,c,t]$, we found that using a random choice in grow/prune/swap/change (GPSC) in one Metropolis-Hasting (MH) step \citep*{chipman1998bayesian}, is not sufficient to grow large trees for our model. This is not a drawback of the proposal mechanism, but is instead primarily due to the notion that following this clustering process would distribute the data entries to many small trees, if any large tree has not yet formed. In other words, the goal is to have our model prioritize ``first in growing the tree, second in clustering" instead of the other order.

Therefore, we devise a new Gibbs sampling scheme, which sequentially samples the full conditional distribution $[s_k|(s_k)]$, $[c_k|(c_k)]$ and $[t_k|(t_k)]$. For each update, MH criterion is used. We restrict updates of $\bf c,t$ that result in an empty node such that $\bf s$ do not change in these steps. The major difference in this approach compared to the GPSC method is that, rather than one random change in one random node, we use micro steps to exhaustively explore possible changes in every node, which increases chain convergence.

Besides increasing the convergence rate, the other function of the Gibbs sampler is to force each change in the tree structure to be small and local. Although some radical change steps \citep*{wu2007bayesian,pratola2013efficient} can facilitate the jumps between the modes in a single tree, for a mixture of trees, local changes and mode sticking are useful to prevent label switching. 

\subsection{Clustering: Updating $[\bf W,Z|T, Y]$}

In this step, we take advantage of the latent uniform variable $\bf U$ in the slice sampler. In order to sample from the joint density $[{\bf U,W,Z} | {\bf T}]$, we use the blocked Gibbs sampling again, in the order of the following marginal densities $[{\bf W}  |{\bf Z,T}]$, $[{\bf U}  |{\bf W, Z,T}]$ and $[{\bf Z}  |{\bf W, U, T}]$.

\begin{equation}
\begin{aligned}
&v_j\sim {Beta}(1+\sum_i\boldsymbol{1}(Z_i=j),\alpha+\sum_i\boldsymbol{1}(Z_i>j))\\
&w_j=v_j\prod_{k<j}(1-v_k)\\
&u_i\sim U(0, w_{Z_i})
\end{aligned}
\end{equation}

During the following update of the assignment ${\bf Z} $, two variables simplify the computation: the truncation effects of the $u_i$'s keep the number of candidate states finite; the values of $\boldsymbol\theta$ in each tree $\bf T$ circumvent the need to recompute the marginal likelihood. The probability of assigning the $i$th observation to the $j$th tree is

\begin{equation}
[Z_i=j|{\bf W, Y, T}_j]\propto \boldsymbol{1}(w_j>u_i)[Y_i|{\bf T}_j]
\end{equation}

%\subsection{Weak Tree Remedy}

%In the above sampling scheme, only the trees with data assigned are updated in each iteration. We further set up a minimum data number for each tree. If the data count of a certain tree falls below it, that tree is not updated further. This is to prevent unnecessary over-pruning in the trees with too little data (weak trees). 

%On the other hand, with this restriction we do observe one caveat: once the cluster number is reduced, it is unlikely to increase. This is due to the lack of updates in any additional tree. To remedy this problem, in each iteration we randomly assign a fixed number of observations to the first weak tree (ordered by $j$), update $T$ for several steps and then revert the assignment. This operation does not affect the likelihood as it is only a proposal step. As the result, in the following update of $Z$, the weak tree is more likely to allocated with data.

\subsection{Posterior Inference: Choosing the Best Ensemble of Trees}
In the posterior analyses of the Markov chain, we could consider the marginal likelihood $\prod_i \int_{Z_i}[y_i|T_{Z_i}]dP(Z_i)$, but it would involve costly allocation of the data over all the candidate trees. Therefore, we use the joint likelihood with the tree assignment $\prod_i[y_i|T_{Z_i}][Z_i]$ as the criterion for choosing the best ensemble of trees.

For prediction purposes, we define two types of estimators: a cluster-specific estimator and an ensemble estimator. The former is defined as $\mathbb{E} (\theta | T_{Z_i})$, where assignment $Z_i$ is known. The latter is defined as $\sum_j w_j \mathbb{E} (\theta | T_{Z_j})$, where assignment $Z_i$ is unknown. In theory, conditional on the collected posterior samples being the same, the cluster-specific estimator has smaller variance than the ensemble estimator. In practice, the latter is more applicable since we often do not know $Z_i$ for a new observation until $Y_i$ is observed.
 
\section{Simulation Studies}
%Previous studies on Bayesian CART  have proposed various methods that involve drastic changes in the shape of a tree, in order to facilitate the move of the chain from one local maximum to another. In BET, however, since the multi-modal issue is already accounted for by its mixing nature, for each mixture component, the limited movement is in fact preferable. Therefore, we use Gibbs sampling to keep the changes as local and small as possible. 

In this section, we demonstrate the clustering capability of BET through simulations of three scenarios: (1) single way of partitioning in $\bf X$ with unimodal $[\bf  Y|X]$; (2) single way of  partitioning in $\bf X$ with multi-modal $[\bf  Y|X]$;  (3) multiple ways of partitioning of $\bf X$ with possible multi-modal $[\bf  Y|X]$. For case (1), the model should converge to one cluster, whereas cases (2) and (3) should both result in multiple clusters.

%As our focus is to cluster the data, instead of comprehensively exploring the tree space, only one of the equivalent tree structures may be discovered; however, this would not interfere with the inference prediction.

\subsection{Unique Partitioning Scheme with Unimodal Likelihood}

This is the simplest yet still likely scenario, in which the data is homogeneous and can be easily classified with a unique partition. We adopt the simulation setting used by \cite*{wu2007bayesian} and generate 300 test samples as shown in Table~\ref{table:sim1_setting}. In this dataset, three partitions can be classified based on combinations of either $\{X_1,X_2\}$ or  $\{X_2,X_3\}$, where two choices are equivalent.

\begin{table}[ht!]
	\begin{center}
		\begin{tabular}{ r  | lll |l}
			\hline                        
			Index&				$X_1$ & $X_2$ & $X_3$ & Y\\
			\hline                        
			1...100 & U(0.1,0.4)&U(0.1,0.4)&  U(0.6,0.9) & N(1.0,$0.5^2$)\\
			101...200& U(0.1,0.4)& U(0.6,0.9)&  U(0.6,0.9)&N(3.0,$0.5^2$)\\
			201...300 &  U(0.6,0.9) & U(0.1,0.9)& U(0.1,0.4)&N(5.0,$0.5^2$)\\
			\hline                        
		\end{tabular}
	\end{center}
	\caption{Setting for Simulation Study I}
	\label{table:sim1_setting}
\end{table}

\begin{figure}[ht!]
	\begin{center}
		\includegraphics[width=.6\columnwidth]{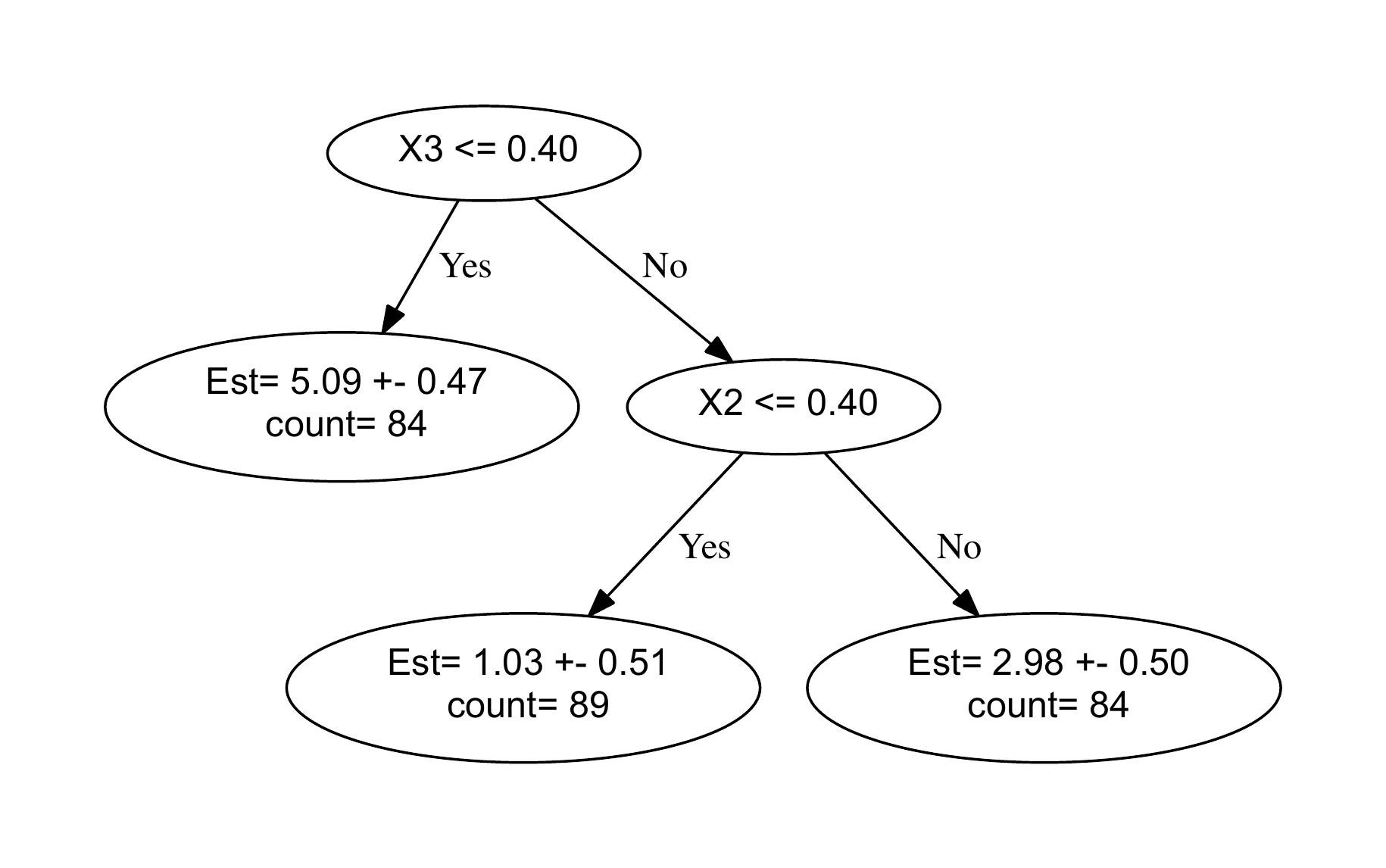}
		\caption{Simulation Study I shows one cluster is found and the partitioning scheme is correctly uncovered}
	\end{center}
	\label{sim1_result}
\end{figure}

We ran the BET model on the data for 10,000 steps and used the last 5,000 steps as the results. The best tree is shown in Figure~1. Three partitions are correctly identified. The choice of $X$ and splitting values are consistent with the simulation parameters. More importantly, the majority of observations are clustered together in a single tree. In each partition, the number of observations and the parameter estimates are very close to the true values. This simulation shows that BET functions as a simple Bayesian CART method when there is no need of clustering.

\subsection{Unique Partitioning Scheme with Multi-modal Likelihood}

This example is to illustrate the scenario of having a``mixture distribution inside one tree".
We duplicate the previous observations in each partition and change one set of their means. As shown in Table~\ref{table:sim2_setting}, each partition now becomes bimodal. Since partitioning based on $\{X_1, X_3\}$ is equivalent to $\{X_1, X_2\}$, we drop $X_3$ for clearer demonstration.

\begin{table}[ht!]
	\begin{center}
		\begin{tabular}{ r  | l l | l }
			\hline                        
			Index&				$X_1$ & $X_2$  & Y\\
			\hline                        
			1...100 & U(0.1,0.4)&U(0.1,0.4)&   N(1.0,$0.5^2$)\\
			101...200& U(0.1,0.4)& U(0.6,0.9)&  N(3.0,$0.5^2$)\\
			201...300 &  U(0.6,0.9) & U(0.1,0.9)& N(5.0,$0.5^2$)\\
			\hline
			301...400 & U(0.1,0.4)&U(0.1,0.4)&   N(1.5,$0.5^2$)\\
			401...500& U(0.1,0.4)& U(0.6,0.9)&  N(5.5,$0.5^2$)\\
		    501...600 &  U(0.6,0.9) & U(0.1,0.9)& N(3.5,$0.5^2$)\\
			\hline                        
		\end{tabular}
	\end{center}
	\caption{Setting for Simulation Study II}
	\label{table:sim2_setting}
\end{table}

\begin{figure}[ht!]

          \begin{subfigure}[b]{.7\columnwidth}
		\includegraphics[width=1\columnwidth]{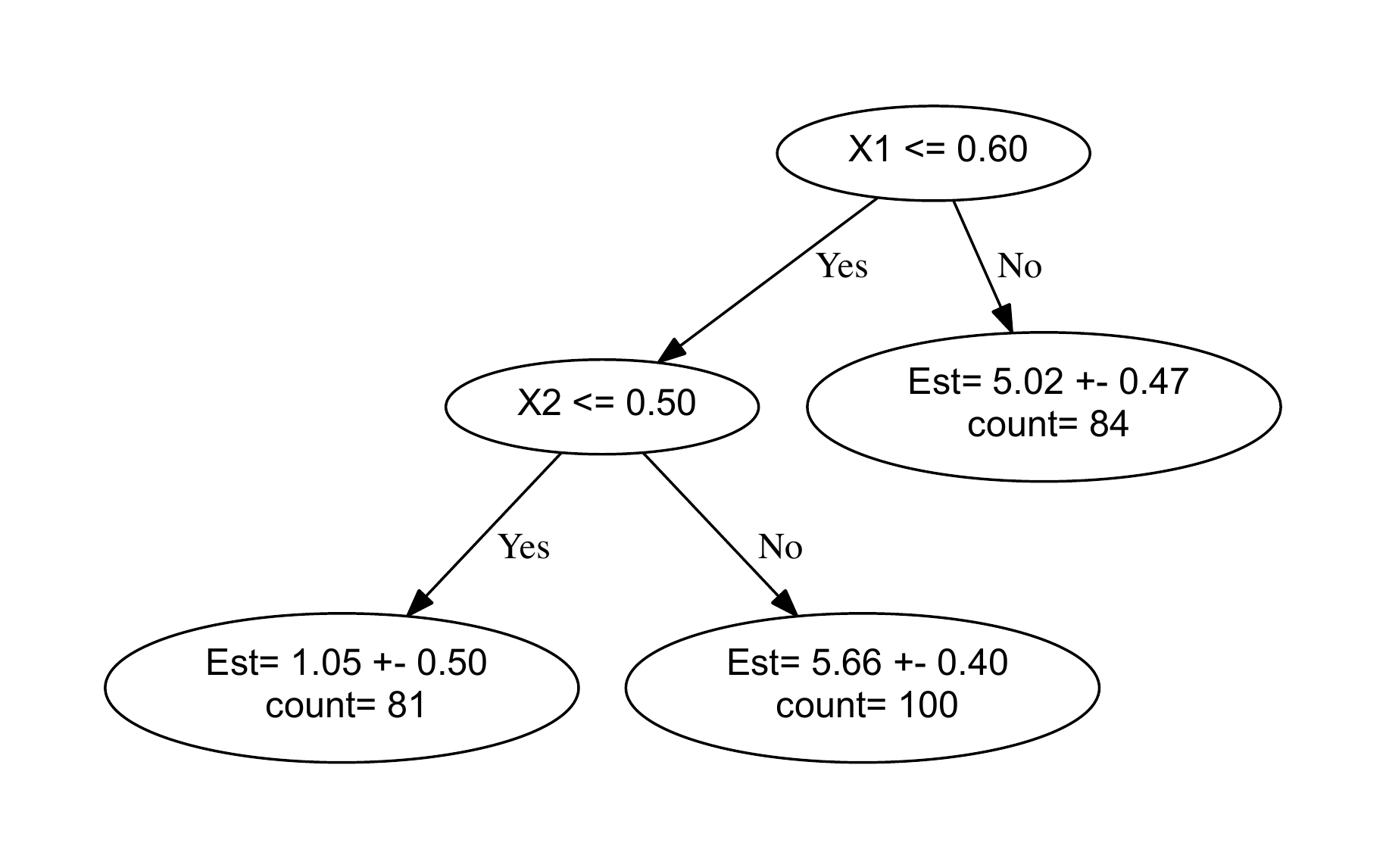}
		\end{subfigure}
		\begin{subfigure}[b]{.7\columnwidth}
			\includegraphics[width=1\columnwidth]{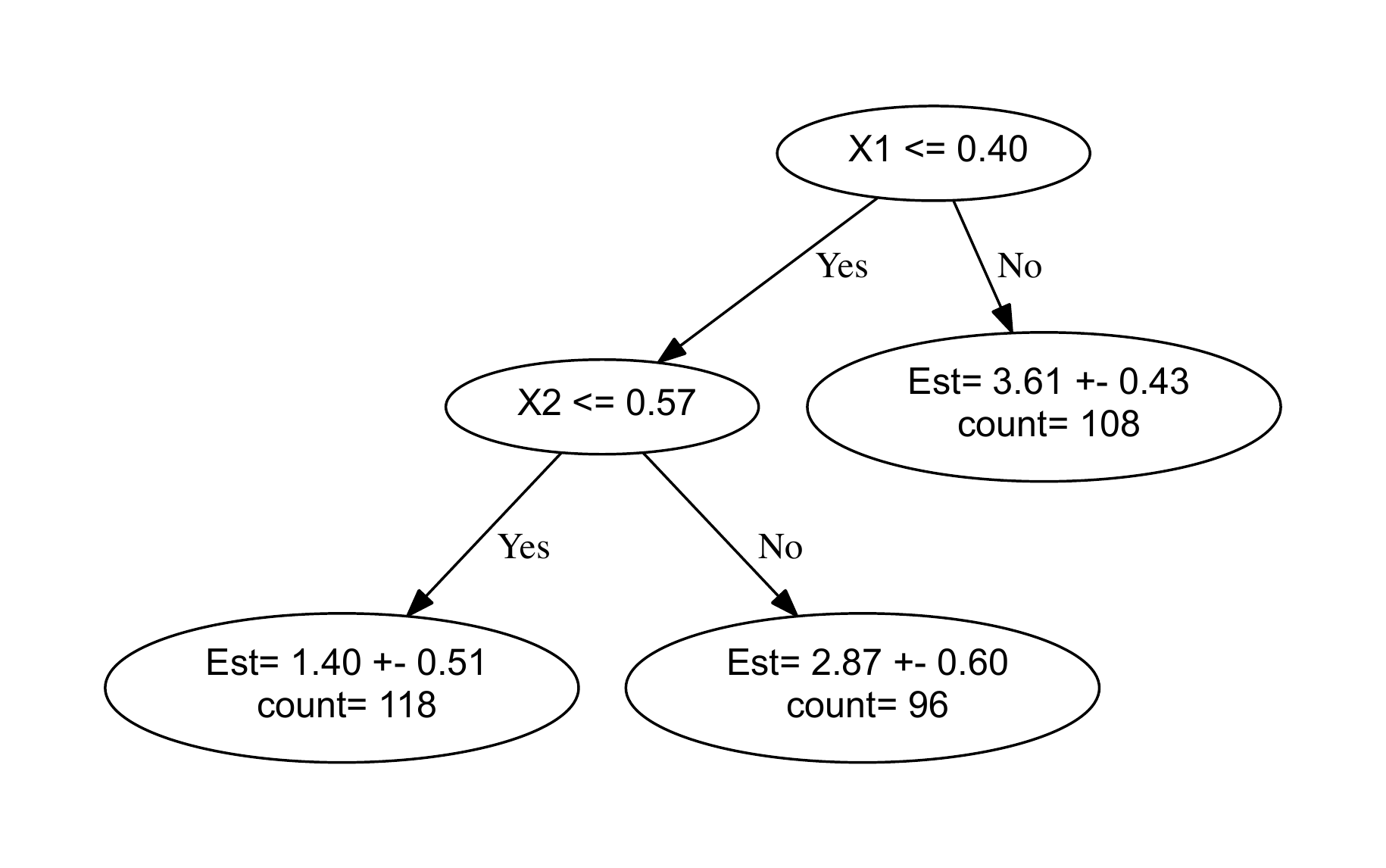}
		\end{subfigure}
				\caption{Simulation Study II shows two clusters are found and parameters are correctly estimated}
	\label{fig:sim2_result}
\end{figure}

The results are shown in Figure~\ref{fig:sim2_result}. Two clusters are correctly identified by BET. The fitted splitting criteria and the estimated parameters are consistent with the true values. It is interesting to note that in the leftmost nodes of the two clusters, there is a significant overlap in distribution (within one standard deviation of normal means). As a result, compared with the original counts in the data generation, some randomness is observed in these two nodes. Nevertheless, the two fitted trees are almost the same as anticipated.

\subsection{Mixed Partitioning Scheme with Unimodal or Multi-modal Likelihood}

\begin{figure}[ht!]
	
	\begin{subfigure}[b]{.4\columnwidth}
		\includegraphics[width=1\columnwidth]{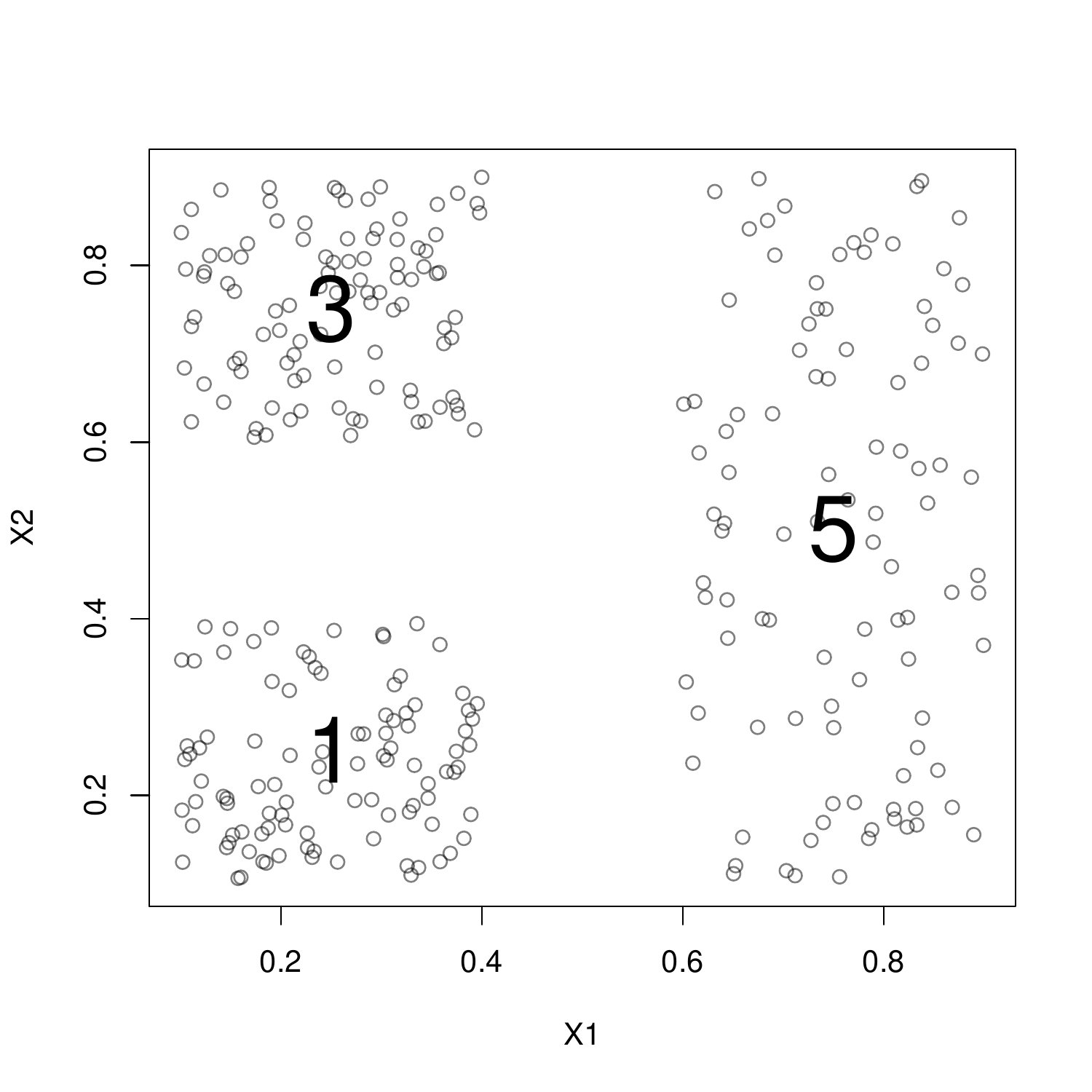}
	\end{subfigure}
	\begin{subfigure}[b]{.4\columnwidth}
		\includegraphics[width=1\columnwidth]{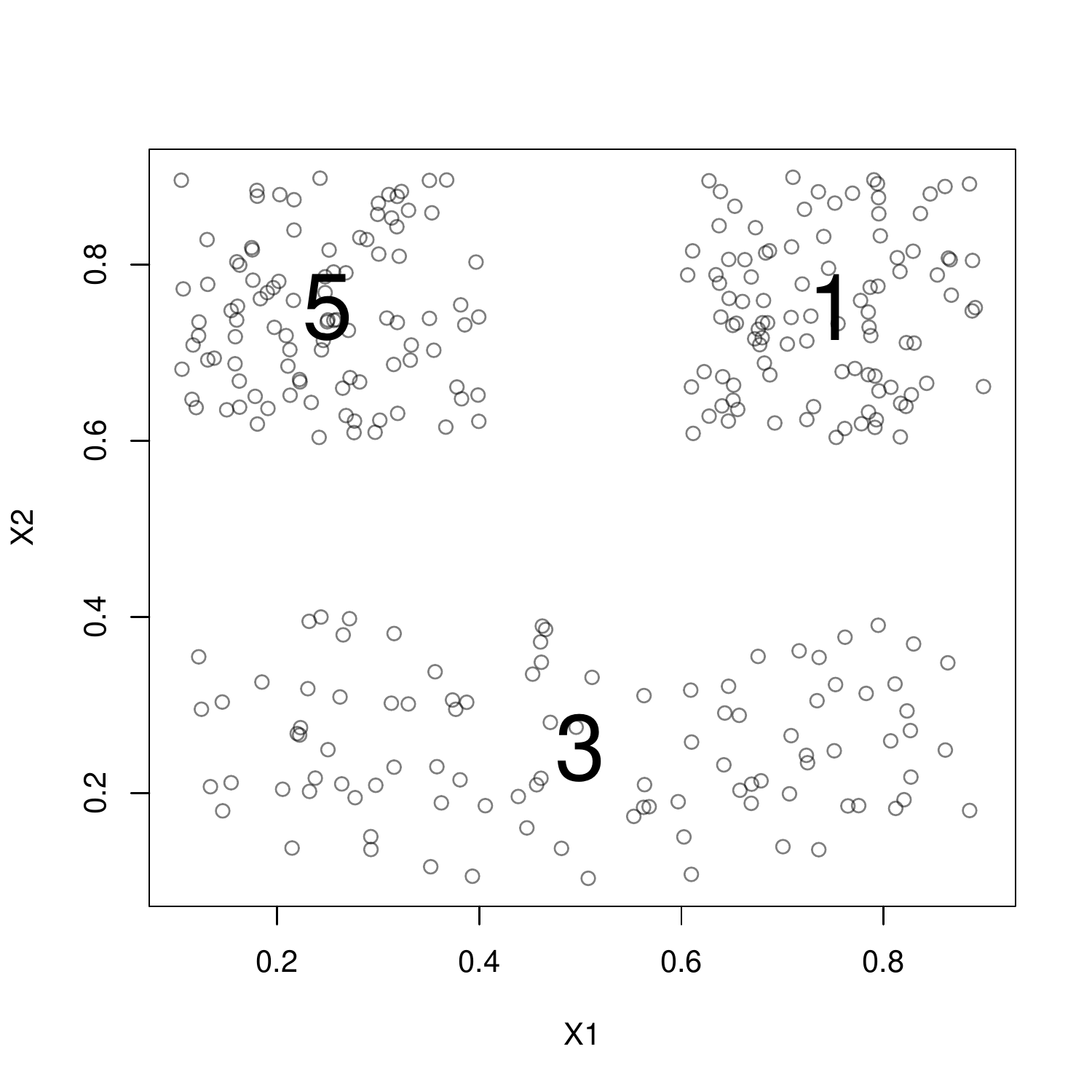}
	\end{subfigure}
	\caption{Simulation Study III has data from different partitioning schemes mixed together. The means are labeled in the center of each region. The shared region in the upper left has mixed means.}
	\label{fig:sim3_diagram}
\end{figure}

This scenario reflects the most complicated case, which is quite realistic in large, heterogeneous data. We again duplicate the data in Table~\ref{table:sim1_setting} and then change the partition scheme for indices $301...600$. This further increases the challenge as shown in Figure~\ref{fig:sim3_diagram}. The data are now formed by a mixed partitioning scheme, which has both overlapping and unique regions. Their distribution means are labeled in each region. It is worth noting that the shared partition in the upper left is bimodal.

\begin{table}[ht!]
	\begin{center}
		\begin{tabular}{ r  | l l |l }
			\hline                        
			Index&				$X_1$ & $X_2$ & Y\\
			\hline                        
			1...100 & U(0.1,0.4)&U(0.1,0.4)&   N(1.0,$0.5^2$)\\
			101...200& U(0.1,0.4)& U(0.6,0.9)& N(3.0,$0.5^2$)\\
			201...300 &  U(0.6,0.9) & U(0.1,0.9)& N(5.0,$0.5^2$)\\
			\hline
			301...400 & U(0.1,0.4)&U(0.6,0.9)&   N(5.0,$0.5^2$)\\
			401...500& U(0.6,0.9)& U(0.6,0.9)&  N(1.0,$0.5^2$)\\
			501...600 &  U(0.1,0.9) & U(0.1,0.4)& N(3.0,$0.5^2$)\\
			\hline                        
		\end{tabular}
	\end{center}
	\caption{Setting for Simulation Study III}
	\label{table:sim3_setting}
\end{table}

To ensure convergence, we ran the model for 20,000 steps and discard the first 10,000 as burn-in steps. Among the total 10,000 steps, the step numbers that correspond to 1,2 and 3 clusters are 214, 9665  and 121. Clearly, the 2-cluster is the most probable model for the data.

%We found some step correspond to a single cluster with a deep tree. This is not surprising since when the data are concentrated in one cluster, their likelihood tends to dominate over the tree prior and results in multiple levels of splitting. As the cluster number increases, the data are scattered and the prior forces the tree to revert to simpler forms. 

\begin{figure}[ht!]
	
	\begin{subfigure}[b]{.8\columnwidth}
		\includegraphics[width=1\columnwidth]{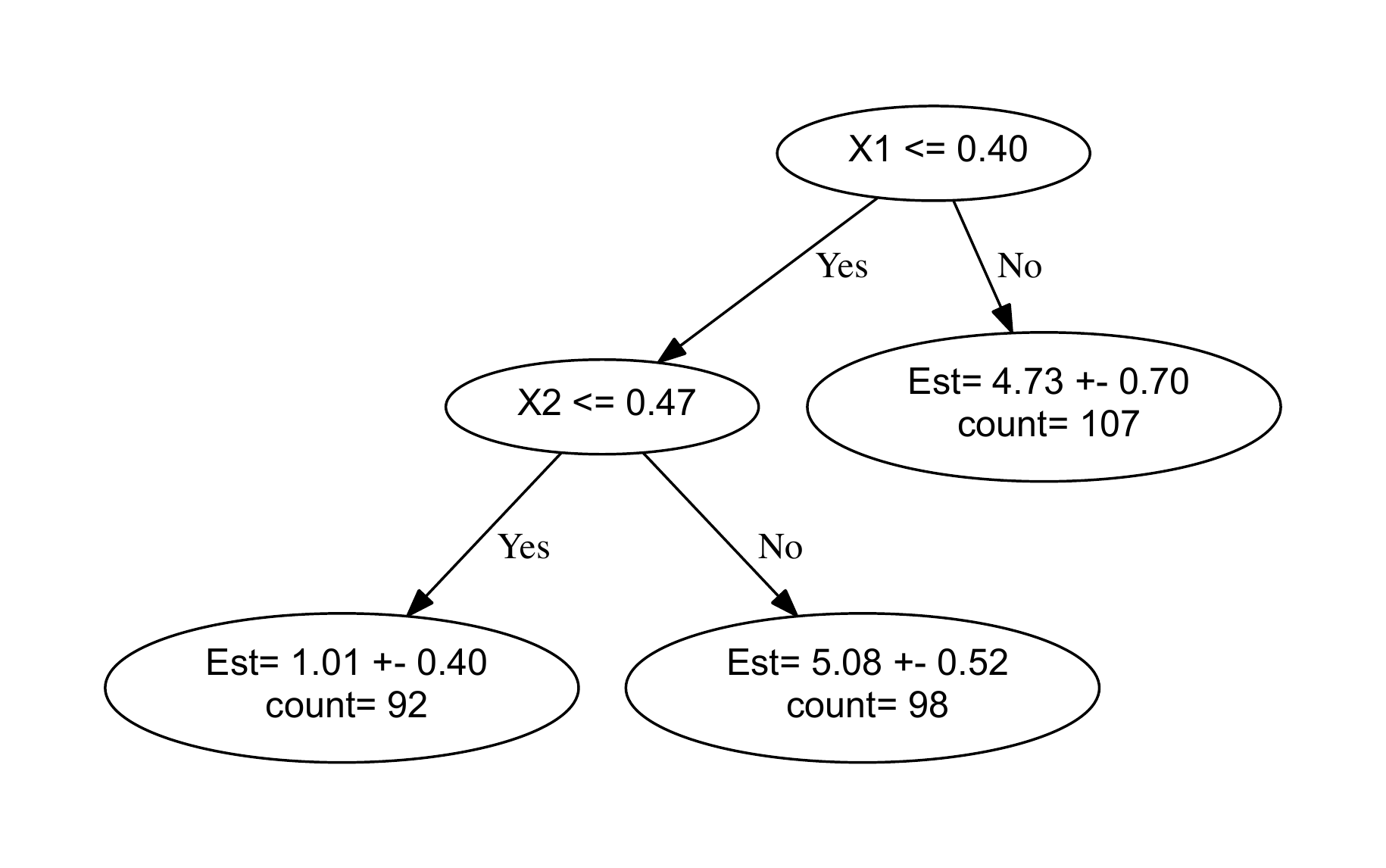}
		\caption{Cluster 1}
	\end{subfigure}
	\begin{subfigure}[b]{.8\columnwidth}
		\includegraphics[width=1\columnwidth]{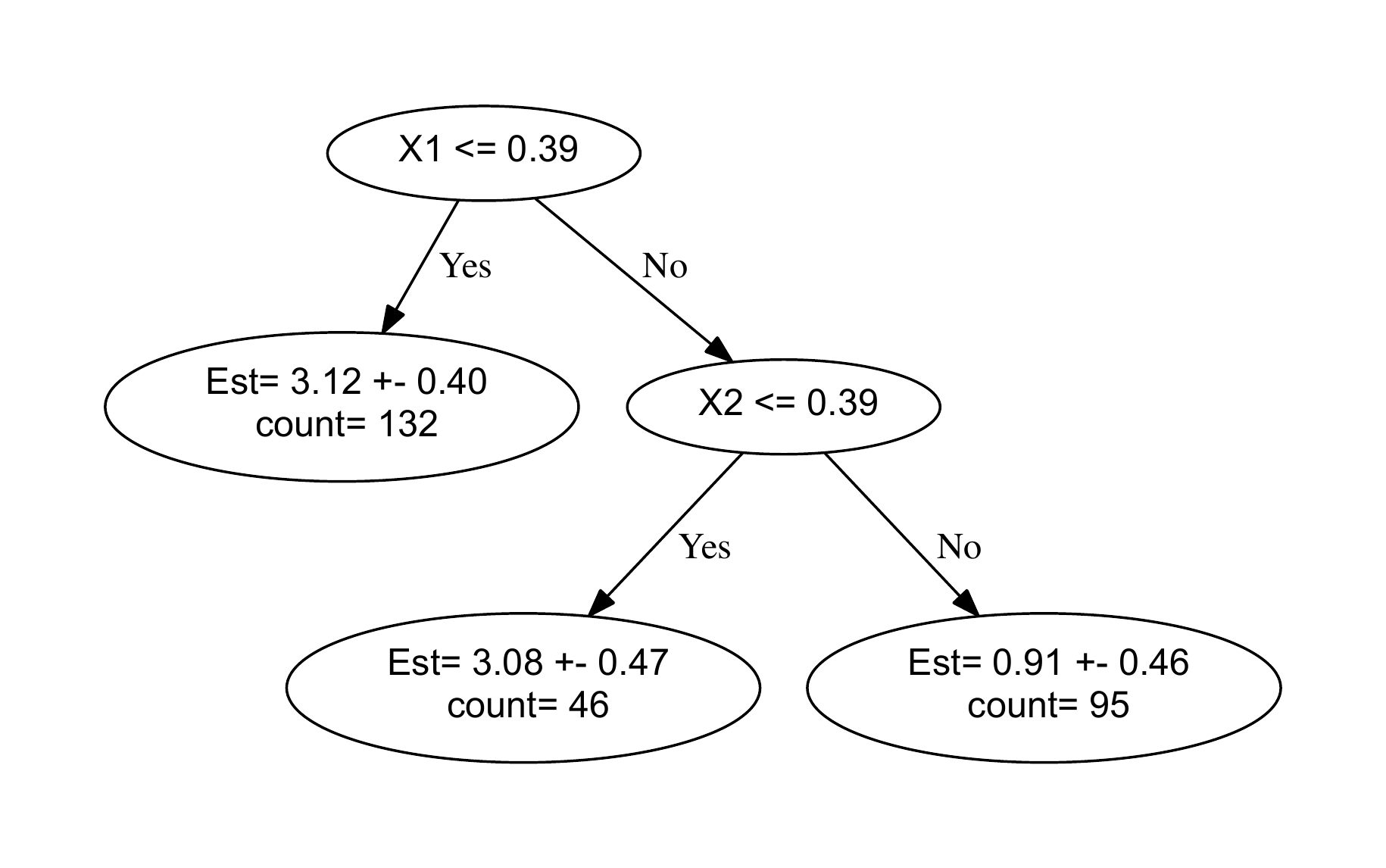}
		\caption{Cluster 2}
	\end{subfigure}
	\caption{Simulation Study III shows two clusters correctly identified}
	\label{fig:sim3_result}
\end{figure}

The best ensemble of trees are shown in Figure~\ref{fig:sim3_result}. This is consistent to the data generation diagram in Figure~\ref{fig:sim3_diagram}, since the two means in the upper left region are exchangeable due to its bimodal nature.

These simulation results show that BET is capable of detecting not only the mixing of distributions, but also the mixing of trees. In the following sections, we test its prediction performance through real data examples.

\section{Breast Cancer Data Example}

We first demonstrate the performance of the Gibbs sampler in the BET model with homogenous data. We examined breast cancer data available from the machine learning repository of the University of California at Irvine \citep*{Bache+Lichman:2013}. This data was originally the work of \cite*{wolberg1990multisurface} and has been subsequently utilized in numerous application studies. We focus on the outcome of breast cancer as the response variable. We define benign and malignant results as $y=0$ and $y=1$, respectively. Our application includes the nine clinical covariates, which are clump thickness, uniformity of cell size, uniformity of cell shape, marginal adhesion, single epithelial cell size, bare nuclei, bland chromatin, normal nucleoli and mitoses.  We consider data for 683 females who have no missing values for the outcome and covariates at hand. We ran the model for 110,000 steps and discarded the first 10,000 steps.

\begin{figure}[ht!]
	\begin{subfigure}[b]{.4\columnwidth}
		\includegraphics[width=1\columnwidth]{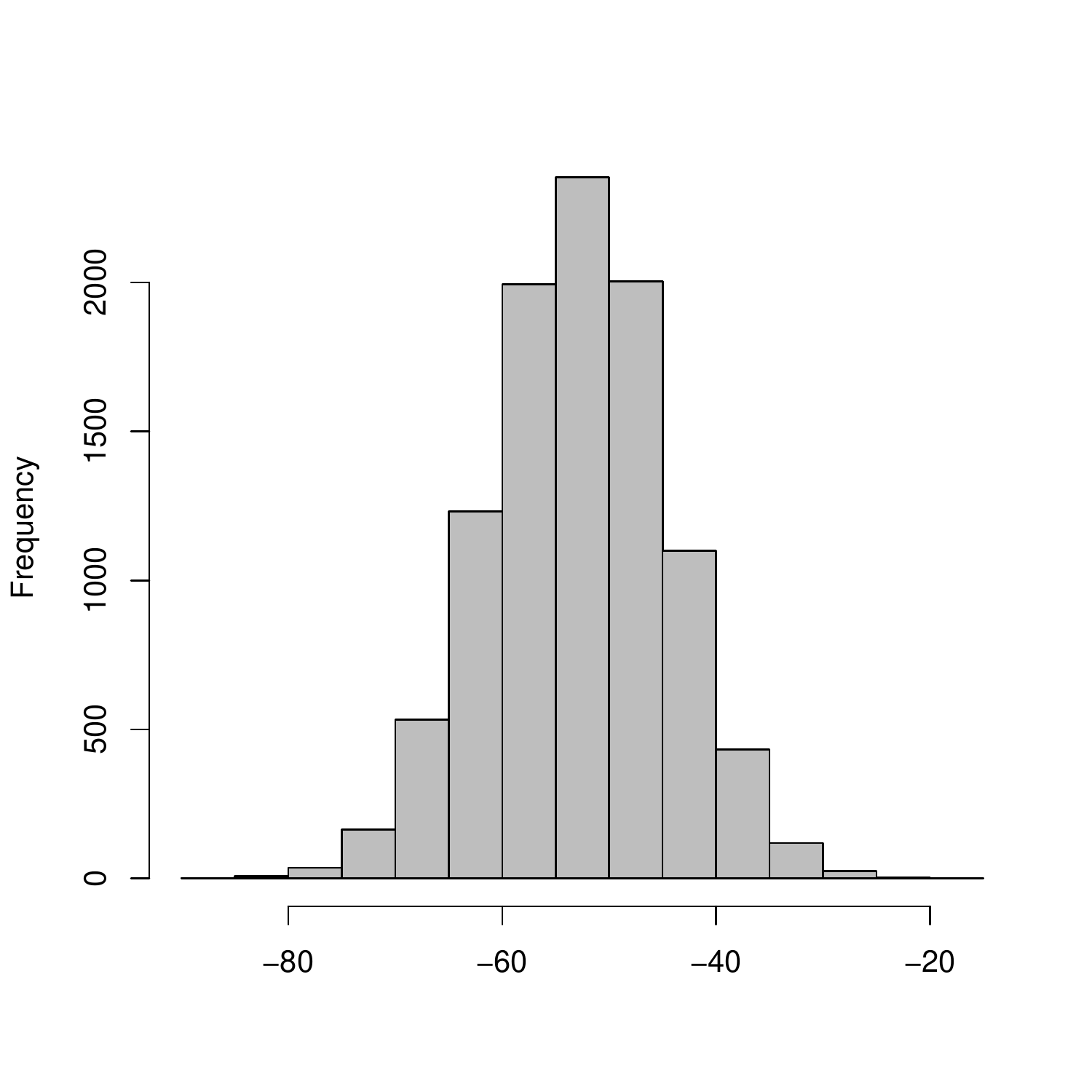}
		\caption{Log-likelihood given assignment}
	\end{subfigure}
	\begin{subfigure}[b]{.4\columnwidth}
		\includegraphics[width=1\columnwidth]{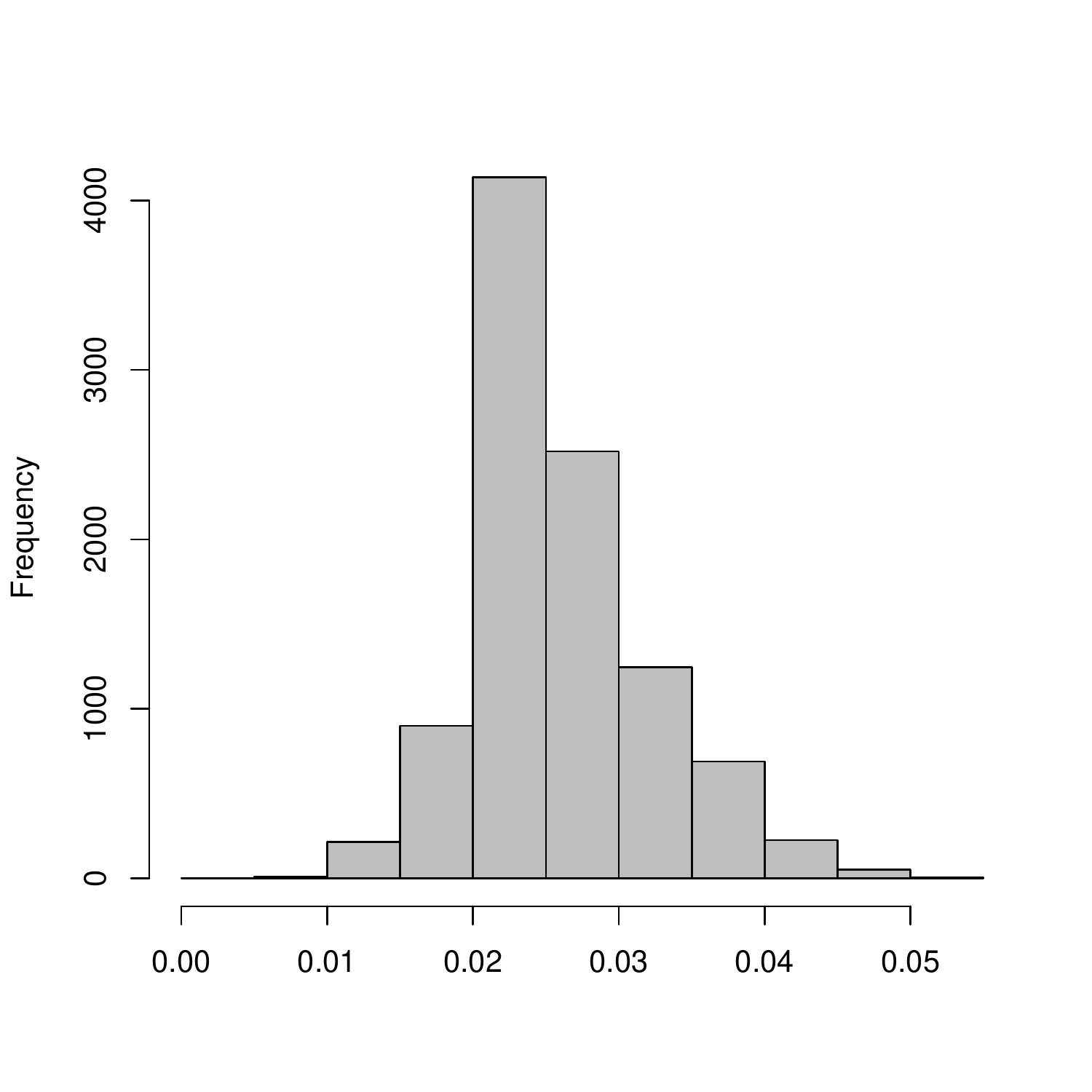}
		\caption{Misclassification rate with cluster-specific estimator}
	\end{subfigure}
		\begin{subfigure}[b]{.8\columnwidth}
			\includegraphics[width=1\columnwidth]{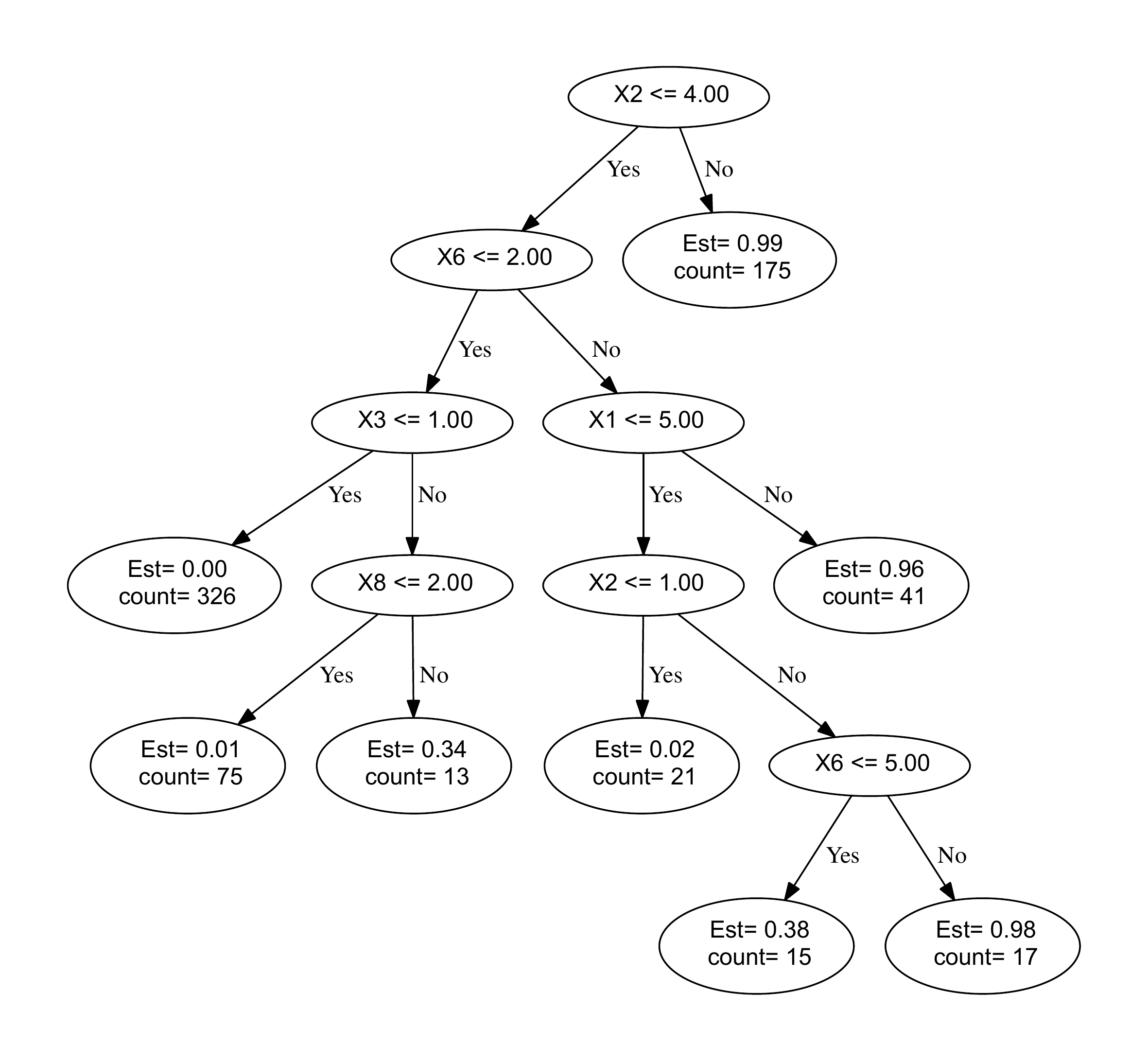}
			\caption{Best ensemble: only one cluster is found}
		\end{subfigure}
	\caption{Result of breast cancer test}
	\label{fig:breast_cancer_results}
\end{figure}

The results show the chain converges to the joint log-likelihood $[Y,Z|T]$  at at mean $-138.3$ and the conditional log-likelihood $[Y|Z,T]$ at mean $-53.8$ (Figure~\ref{fig:breast_cancer_results}(a)). The second number is slightly improved over the result reported by \cite*{wu2007bayesian}.

We use $0.5$ as the cut-off point of estimates to dichotomize 0 and 1, then use the proportion of incorrect classification as the misclassification rate (MCR). The majority of the collected steps correspond to only one cluster (Figure~\ref{fig:breast_cancer_results}(c)). This suggests the data is highly homogeneous, which contributes to the low misclassification rate (Figure~\ref{fig:breast_cancer_results}(b)).  The MCR is $0.025\pm0.007$ (Figure~\ref{fig:breast_cancer_results}(c)), with the minimum equal to $0.013$.

To assess prediction performance, we carried out half-split cross-validation.  We randomly selected 342 observations as the training set. Predictions are made on the remaining 341 observations. We use the ensemble estimator for the MCR calculation. We repeated the test for ten separate times, each with a different seed for the random split. The resulting MCR is as low as $0.036\pm 0.006$, which is close to the one reported by \cite*{wu2007bayesian}. These results support that the Gibbs sampling algorithm can rapidly move the chain in the tree structure space ${\mathcal{T}}$.

Next we test BET on a large dataset, which possibly contains heterogeneous data and simultaneously illustrate its regression performance.

\section{Cystic Fibrosis Data Example}

We used lung function data obtained from the Cystic Fibrosis Foundation Patent Registry \citep*{cf2012}. Percent predicted of forced expiratory volume in 1 second (FEV$_1\%$) is a continuous measure of lung function in cystic fibrosis (CF) patients obtained at each clinical visit. We have previously demonstrated that the rates of FEV$_1\%$ change nonlinearly \citep*{szczesniak2013semiparametric}, which can be described via semiparametric regression using penalized cubic splines. Although trees may not outperform spline methods in prediction of continuous outcomes, they provide reliable information for variable selection when traditional methods such as p-value inspection and AIC may fall short.

We used longitudinal data from 3,500 subjects (a total of 60,000 entries) from the CF dataset and utilized eight clinical covariates: baseline FEV$_1\%$, age, gender, infections (each abbreviated as MRSA, PA, BC, CFRD) and insurance (a measure of socioeconomic status, SES). We randomly selected longitudinal data from 2,943 subjects (roughly 50,000 entries) and used these data as the training set. We then carried out prediction on the remaining 10,000 entries.

\begin{figure}[ht!]
	\begin{subfigure}[b]{.5\columnwidth}
		\includegraphics[width=1\columnwidth]{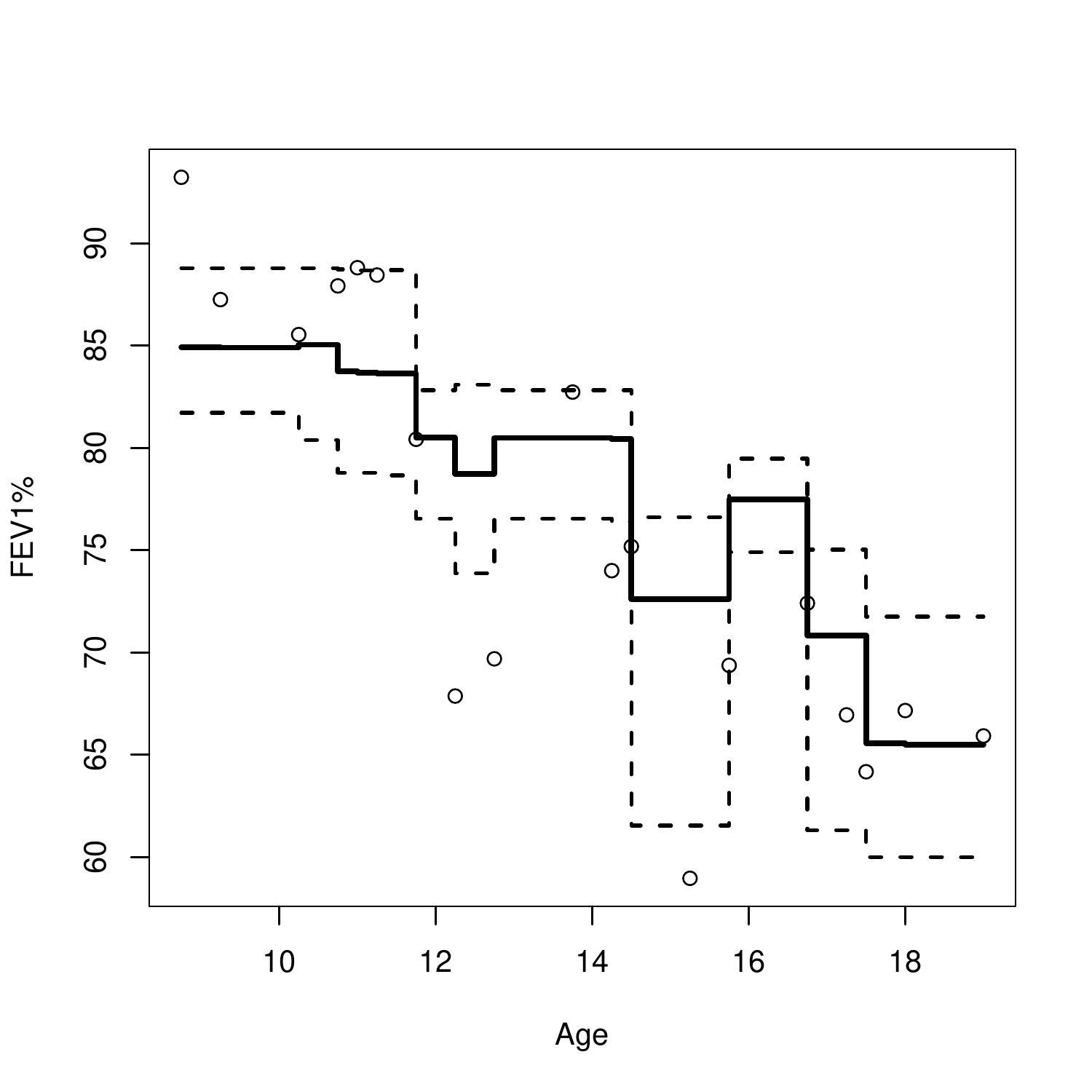}
		\caption{True values (dots) and the 95\% credible interval of the BET predicted (solid) in a single subject.}
	\end{subfigure}
	\begin{subfigure}[b]{.5\columnwidth}
		\includegraphics[width=1\columnwidth]{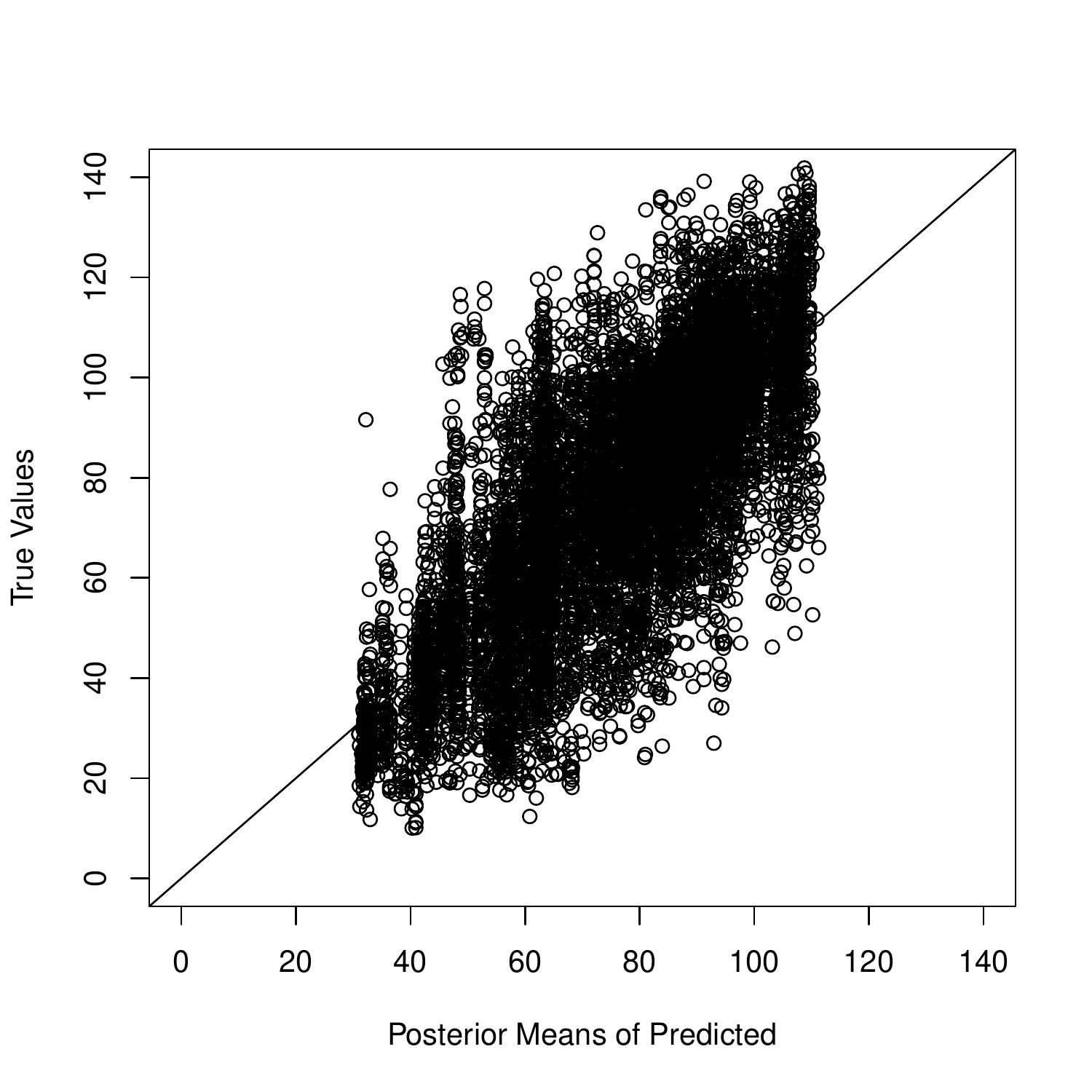}
		\caption{True values and the BET predicted means: the estimates are piecewise constant and seem unbiased.}
	\end{subfigure}
		\begin{subfigure}[b]{.5\columnwidth}
			\includegraphics[width=1\columnwidth]{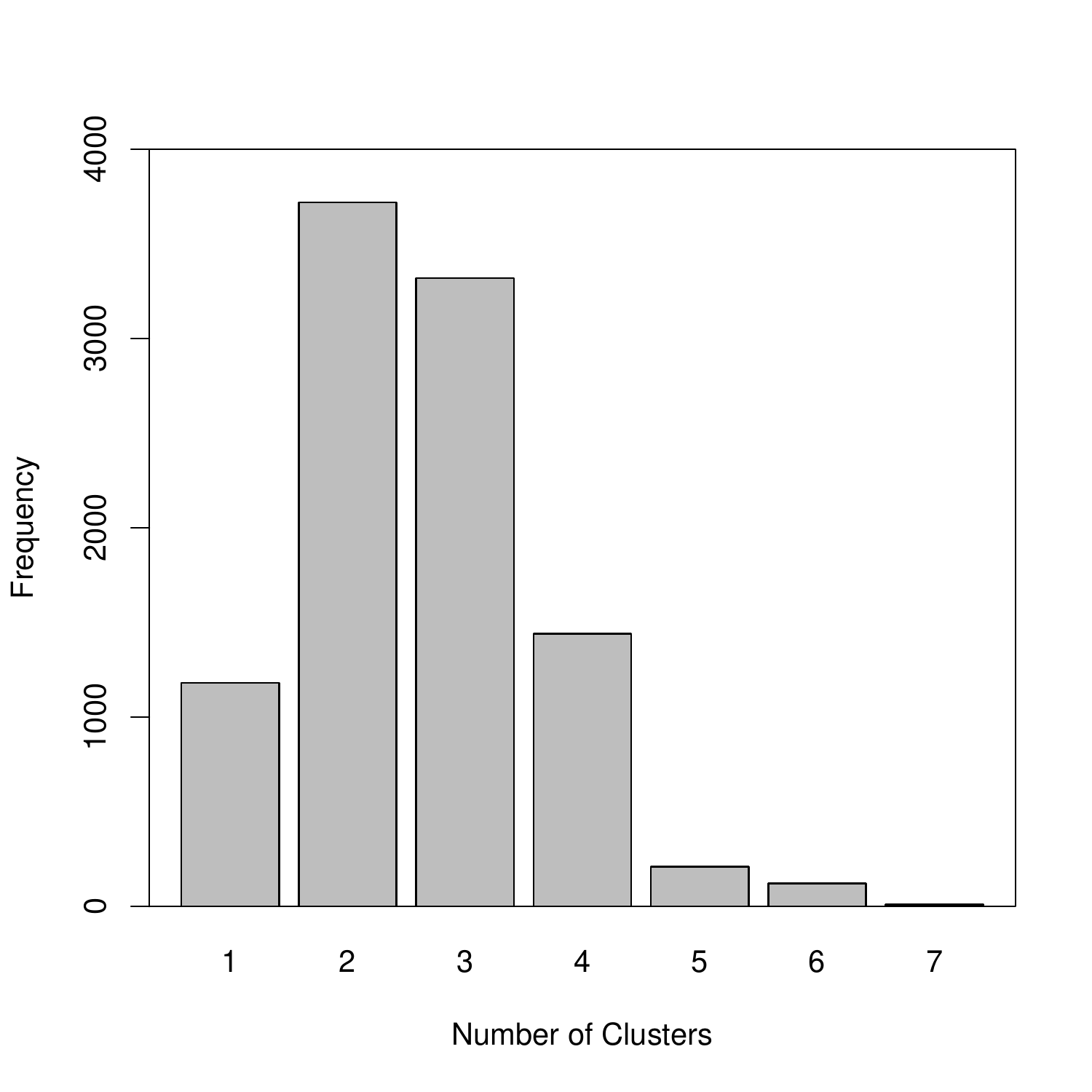}
			\caption{Number of clusters found}
		\end{subfigure}
			\begin{subfigure}[b]{.5\columnwidth}
				\includegraphics[width=1\columnwidth]{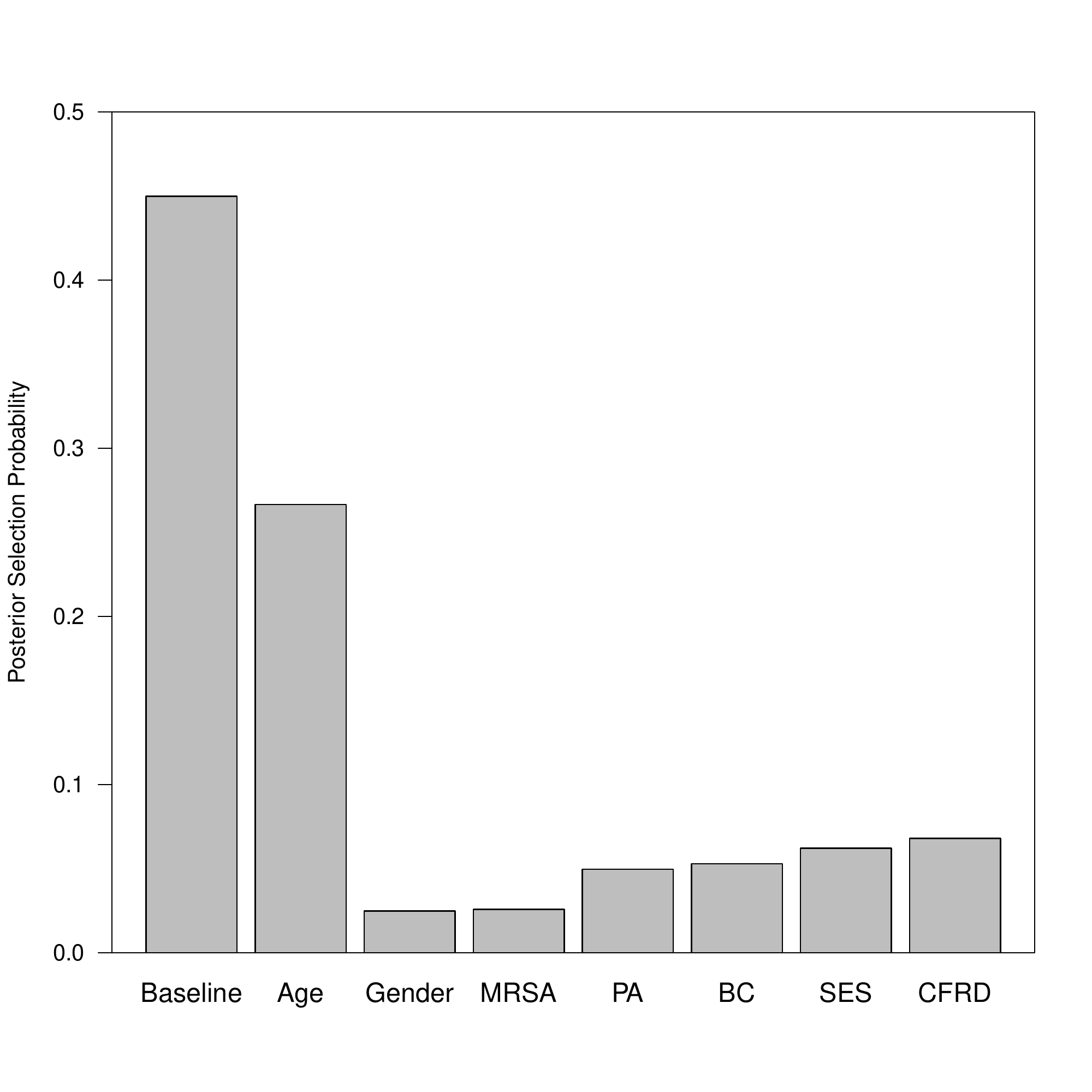}
				\caption{The variable selection probabilities}
			\end{subfigure}
	\caption{Results of FEV$_1\%$ fitting and prediction.}
	\label{fig:cf_results}
\end{figure}

We illustrate the prediction results of BET in Figure~\ref{fig:cf_results}(a and b). With the assumption of one constant mean per partition, the predicted line takes the shape of step functions, which correctly captured the declining trend of FEV$_1\%$. The prediction seems unbiased, as the difference between the predicted and true values are symmetric about the diagonal line.  We also computed the difference metrics with the true values. The results are listed in Table~\ref{table:cf_results}.

Besides the BET model, we also tested other popular regression tree methods (with corresponding R packages), such as CART ( ``rpart", \cite*{therneau1997introduction}), Random Forests (``randomForest'' \cite*{breiman2001random}),  Boosting (``gbm'' \cite*{friedman2001greedy}) and BART(``bartMachine'' \cite*{chipman2010bart}). Since the tested data are essentially longitudinal data, we can choose whether to group observations by subjects or by entries alone. In the subject-clustered version, we first used one outcome entry of each subject in the prediction subset, computed the most likely cluster and then computed the cluster-specific predictor; in the entry-clustered version, we simply used the ensemble predictor. We did not see an obvious difference between the two predictors. This is likely due to entry-clustered BET achieving better fit, which compensates for less accuracy of the ensemble predictor.

\begin{table}[ht!]
	\begin{center}
		\begin{tabular}{ l  | l l l |l}
			\hline                        
			Model&				RMSE & MAD\\
			\hline
			Spline Regression& 16.60 & 10.07 \\
			\hline
			CART& 18.07& 10.32\\
			Random Forests (5 trees)&  17.38& 11.29\\
			Random Forests (50 trees)&  17.13& 11.28\\
			Boosting (1000 trees) &20.34 & 14.22\\
			BART (50 trees) &16.72& 10.32\\
			\hline
			BET  (clustered by subjects) & 16.97 & 10.57\\
			BET  (clustered by entries) & 16.70 & 10.13\\                    
			\hline 
		\end{tabular}
	\end{center}
	\caption{Cross-validation results with various methods applied on cystic fibrosis data.}
	\label{table:cf_results}
\end{table}

In the comparison among the listed tree-based methods, the two Bayesian methods BET and BART provide the closest results to spline regression in prediction accuracy. Similar to the relation between BART and Boosting, BET can be viewed as the Bayes counterpart of Random Forests. Besides the use of prior, one important distinction is that the Random Forests approach uses the average of bootstrap sample estimates, whereas BET uses the weighted average of cluster sample estimates. In Random Forests, the number of bootstrap sample needs to be specified by the user; while in BET, it is determined by the data through Dirichlet process. During this test, Random Forests used 50 trees; while BET converged to only 2-3 trees (Figure~\ref{fig:cf_results}(c)) and achieved similar prediction accuracy. The tree structures are shown in the appendix.

Lastly, we focus on the variable selection issue. Although some information criteria have been established in this area, such as AIC, BIC and Bayes Factor, their mechanisms are complicated by the necessity to fit different models to the data for multiple times. The inclusion probability (reviewed by \cite*{o2009review}) is an attractive alternative, which penalizes the addition of more variables through the inclusion prior. In the tree-based method, however, since multiplicity is not a concern, it is possible to compare several variables of similar importances at the same time, without inclusion or exclusion.

Since multiple trees are present, we use the weighted posterior $\bar\xi=\sum_j w_j \xi_j$ as the measure for variable importance. We plot the variable ranking probability $\bar\xi$ for each covariate (Figure~\ref{fig:cf_results}). The interpretation of this value follows naturally as how likely a covariate is chosen in forming the trees. Therefore, the ranking of $\bar\xi$  reveals the order of importance of the covariates. This concept is quite similar to the variable importance measure invented in Random Forests. The difference is that their index is ranked in the decrease of accuracy after permuting of a covariate; ours is purely a  probabilistic measure. Regardless of this difference, the ranking of variables from the two models are remarkably similar (see Supplementary Information for results of Random Forests): baseline FEV$_1 \%$ and age are the two most important variables while gender and MRSA seem to play the least important roles.

\section{ Discussion}

Empirically, compared with a single model, a group of models (an ensemble) usually has better performance in inference and prediction. This view is also supported in the field of the decision trees. As reviewed by \cite*{hastie2009elements}, the use of decision trees has benefited from multiple-tree methods such as Random Forests, Boosting and Bayesian tree averaging. One interesting question that remains, however, is how many models are enough, or, how many are more than enough?

To address this issue, machine learning algorithms usually resort to repeated tests of cross validation or out-of-bootstrap error calculation, in which ensemble size is gradually increased until performance starts to degrade. In Bayesian model averaging, the number of steps to keep is often ``the more, the better", as long as the steps have low autocorrelation.

Our proposed method, BET, demonstrates a more efficient way to create an ensemble with the same predictive capability but much smaller size. With the help of the Dirichlet process, the self-reinforcing behavior of large clusters reduces the number of needed clusters (sub-models). Rather than using a simple average over many trees, we showed that using a weighted average over a few important trees can provide the same accuracy.

It is worth comparing BET with other mixture models. In the latter, the component distributions are typically continuous and unimodal; in BET, each component tree is discrete, and more importantly, multi-modal itself. This construction could have created caveats in model fitting, as one can imagine only obtaining a large ensemble of very small trees. We circumvented this issue by applying Gibbs sampling in each tree, which rapidly increases the fit of tree to the data during tree growing, and decreases the chance that they are scattered to more clusters. 

It is also of interest to develop an empirical algorithm for BET. One possible extension is to use a local optimization technique (also known as ``greedy algorithm'') under some randomness to explore the tree structure. This implementation may be not difficult, since users can access existing CART packages to grow trees for subsets of data, then update clustering as mentioned previously.

\section*{Acknowledgments}
Partial funding was provided by the Cystic Fibrosis Foundation Research and Development Program (grant number R457-CR11). The authors are grateful to the Cystic Fibrosis Foundation Patient Registry Committee for their thoughtful comments and data dispensation. The breast cancer domain was obtained from the University Medical Centre, Institute of Oncology, Ljubljana, Yugoslavia. The authors thank M. Zwitter and M. Soklic for availability of this data. 

\newpage

\bibliography{reference}

\begin{thebibliography}{20}
\providecommand{\natexlab}[1]{#1}
\providecommand{\url}[1]{\texttt{#1}}
\expandafter\ifx\csname urlstyle\endcsname\relax
  \providecommand{\doi}[1]{doi: #1}\else
  \providecommand{\doi}{doi: \begingroup \urlstyle{rm}\Url}\fi

\bibitem[Bache and Lichman(2013)]{Bache+Lichman:2013}
K.~Bache and M.~Lichman.
\newblock {UCI} machine learning repository, 2013.
\newblock URL \url{http://archive.ics.uci.edu/ml}.

\bibitem[Breiman et~al.(1984)Breiman, Friedman, Olshen, and Stone]{cart84}
L.~Breiman, J.~Friedman, R.~Olshen, and C.~Stone.
\newblock \emph{{Classification and Regression Trees}}.
\newblock Wadsworth and Brooks, Monterey, CA, 1984.

\bibitem[Breiman(2001)]{breiman2001random}
Leo Breiman.
\newblock Random forests.
\newblock \emph{Machine learning}, 45\penalty0 (1):\penalty0 5--32, 2001.

\bibitem[Chipman et~al.(1998)Chipman, George, and
  McCulloch]{chipman1998bayesian}
Hugh~A Chipman, Edward~I George, and Robert~E McCulloch.
\newblock Bayesian cart model search.
\newblock \emph{Journal of the American Statistical Association}, 93\penalty0
  (443):\penalty0 935--948, 1998.

\bibitem[Chipman et~al.(2010)Chipman, George, McCulloch,
  et~al.]{chipman2010bart}
Hugh~A Chipman, Edward~I George, Robert~E McCulloch, et~al.
\newblock Bart: Bayesian additive regression trees.
\newblock \emph{The Annals of Applied Statistics}, 4\penalty0 (1):\penalty0
  266--298, 2010.

\bibitem[Cystic Fibrosis~Foundation(2012)]{cf2012}
1~Cystic Fibrosis~Foundation.
\newblock Cystic fibrosis foundation patient registry 2012 annual data report.
\newblock 2012.

\bibitem[Denison et~al.(1998)Denison, Mallick, and Smith]{denison1998bayesian}
David~GT Denison, Bani~K Mallick, and Adrian~FM Smith.
\newblock A bayesian cart algorithm.
\newblock \emph{Biometrika}, 85\penalty0 (2):\penalty0 363--377, 1998.

\bibitem[Friedman(2001)]{friedman2001greedy}
Jerome~H Friedman.
\newblock Greedy function approximation: a gradient boosting machine.
\newblock \emph{Annals of Statistics}, pages 1189--1232, 2001.

\bibitem[Friedman(2002)]{friedman2002stochastic}
Jerome~H Friedman.
\newblock Stochastic gradient boosting.
\newblock \emph{Computational Statistics \& Data Analysis}, 38\penalty0
  (4):\penalty0 367--378, 2002.

\bibitem[Green and Richardson(2001)]{green2001modelling}
Peter~J Green and Sylvia Richardson.
\newblock Modelling heterogeneity with and without the dirichlet process.
\newblock \emph{Scandinavian journal of statistics}, 28\penalty0 (2):\penalty0
  355--375, 2001.

\bibitem[Hastie et~al.(2009)Hastie, Tibshirani, Friedman, Hastie, Friedman, and
  Tibshirani]{hastie2009elements}
Trevor Hastie, Robert Tibshirani, Jerome Friedman, T~Hastie, J~Friedman, and
  R~Tibshirani.
\newblock \emph{The elements of statistical learning}, volume~2.
\newblock Springer, 2009.

\bibitem[Ishwaran and James(2001)]{ishwaran01}
Hemant Ishwaran and Lancelot~F. James.
\newblock Gibbs sampling methods for stick-breaking priors.
\newblock \emph{Journal of the American Statistical Association}, 96\penalty0
  (453):\penalty0 161--173, 2001.

\bibitem[Kalli et~al.(2011)Kalli, Griffin, and Walker]{kalli2011slice}
Maria Kalli, Jim~E Griffin, and Stephen~G Walker.
\newblock Slice sampling mixture models.
\newblock \emph{Statistics and computing}, 21\penalty0 (1):\penalty0 93--105,
  2011.

\bibitem[Neal(2000)]{neal2000markov}
Radford~M Neal.
\newblock Markov chain sampling methods for dirichlet process mixture models.
\newblock \emph{Journal of computational and graphical statistics}, 9\penalty0
  (2):\penalty0 249--265, 2000.

\bibitem[O'Hara et~al.(2009)O'Hara, Sillanp{\"a}{\"a}, et~al.]{o2009review}
Robert~B O'Hara, Mikko~J Sillanp{\"a}{\"a}, et~al.
\newblock A review of bayesian variable selection methods: what, how and which.
\newblock \emph{Bayesian analysis}, 4\penalty0 (1):\penalty0 85--117, 2009.

\bibitem[Pratola(2013)]{pratola2013efficient}
MT~Pratola.
\newblock Efficient metropolis-hastings proposal mechanisms for bayesian
  regression tree models.
\newblock \emph{arXiv preprint arXiv:1312.1895}, 2013.

\bibitem[Szczesniak et~al.(2013)Szczesniak, McPhail, Duan, Macaluso, Amin, and
  Clancy]{szczesniak2013semiparametric}
Rhonda~D Szczesniak, Gary~L McPhail, Leo~L Duan, Maurizio Macaluso, Raouf~S
  Amin, and John~P Clancy.
\newblock A semiparametric approach to estimate rapid lung function decline in
  cystic fibrosis.
\newblock \emph{Annals of epidemiology}, 23\penalty0 (12):\penalty0 771--777,
  2013.

\bibitem[Therneau et~al.(1997)Therneau, Atkinson,
  et~al.]{therneau1997introduction}
Terry~M Therneau, Elizabeth~J Atkinson, et~al.
\newblock An introduction to recursive partitioning using the rpart routines.
\newblock 1997.

\bibitem[Wolberg and Mangasarian(1990)]{wolberg1990multisurface}
William~H Wolberg and Olvi~L Mangasarian.
\newblock Multisurface method of pattern separation for medical diagnosis
  applied to breast cytology.
\newblock \emph{Proceedings of the national academy of sciences}, 87\penalty0
  (23):\penalty0 9193--9196, 1990.

\bibitem[Wu et~al.(2007)Wu, Tjelmeland, and West]{wu2007bayesian}
Yuhong Wu, H{\aa}kon Tjelmeland, and Mike West.
\newblock Bayesian cart: Prior specification and posterior simulation.
\newblock \emph{Journal of Computational and Graphical Statistics}, 16\penalty0
  (1):\penalty0 44--66, 2007.

\end{thebibliography}
\bibliographystyle{plainnat}
\newcommand{\Appendix}{\appendix\def\thesection{Appendix~\Alph{section}}
\def\thesubsection{\Alph{section}.\arabic{subsection}}
}

\newpage
\section*{Supplementary Information}
Variable Importance calculated by Random Forests with 50 trees, using cystic fibrosis data.
\begin{appendix}
	
\begin{figure}[h!]
		\includegraphics[width=.5\columnwidth]{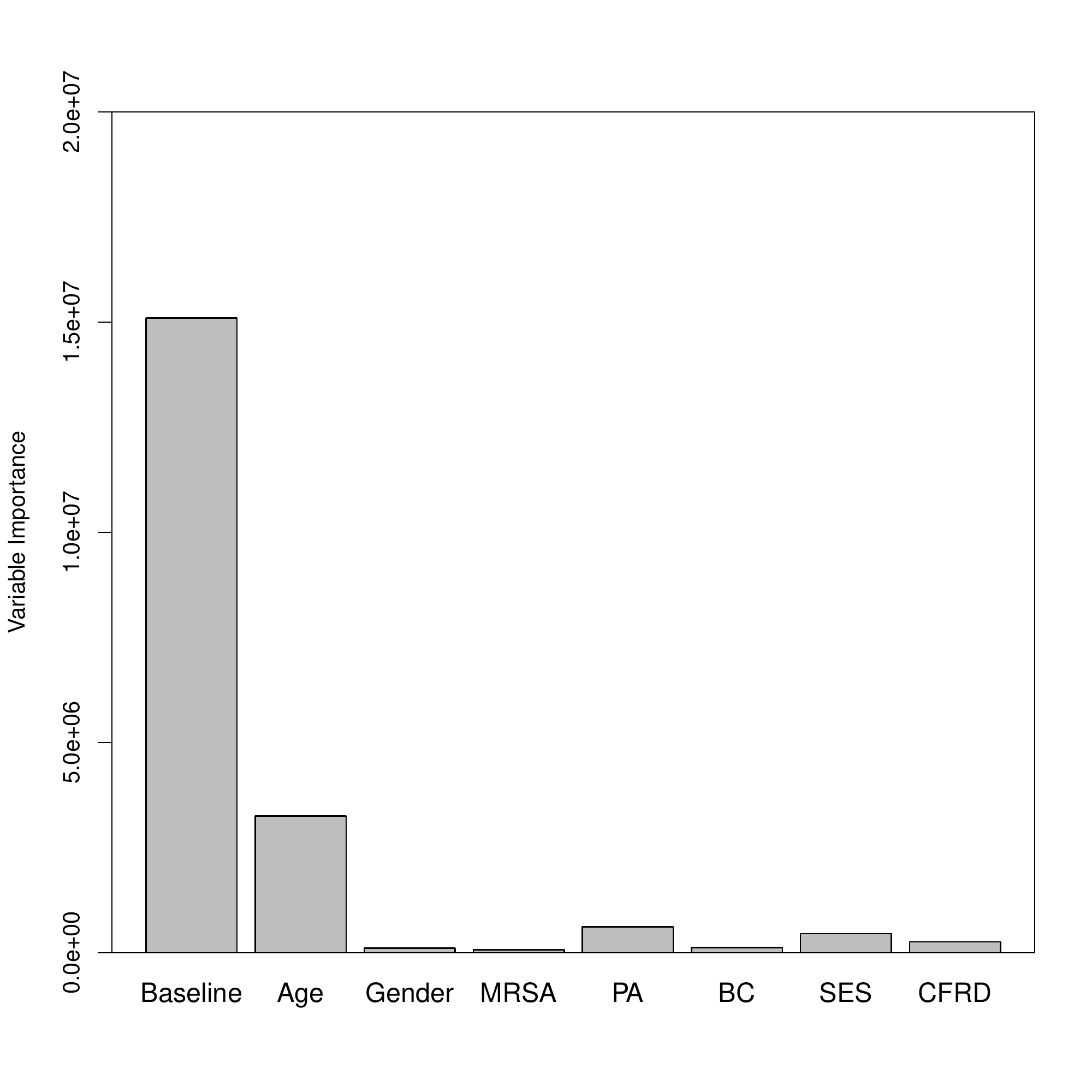}
		\caption{Variable importance calculated by Random Forests with 50 trees}
	\end{figure}

\begin{figure}
	\includegraphics[width=1.3\columnwidth, angle =90]{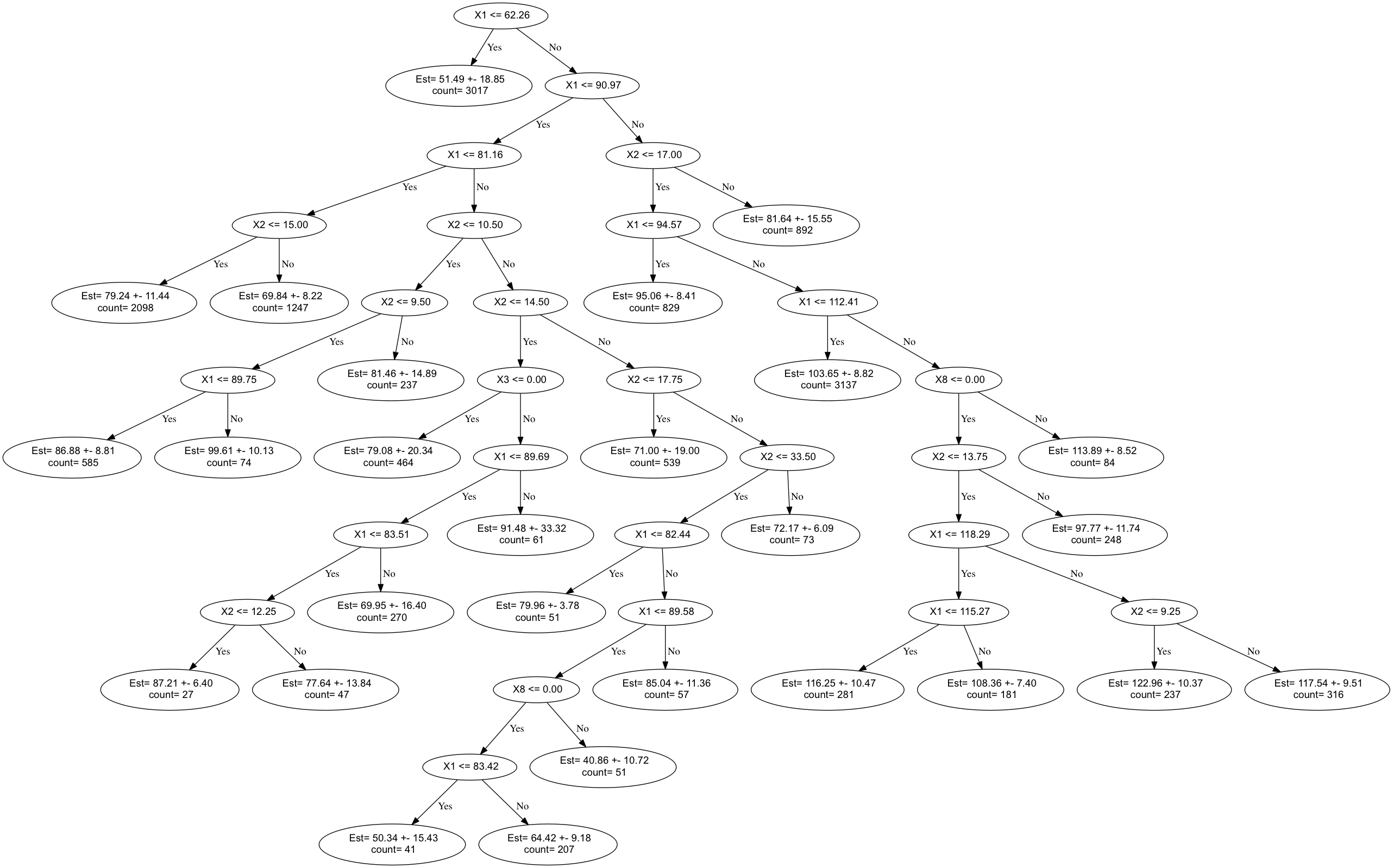}
	\caption{Cluster 1 found by the BET model applied on the cystic fibrosis data}
\end{figure}

\begin{figure}
	\includegraphics[width=1.3\columnwidth, angle =90]{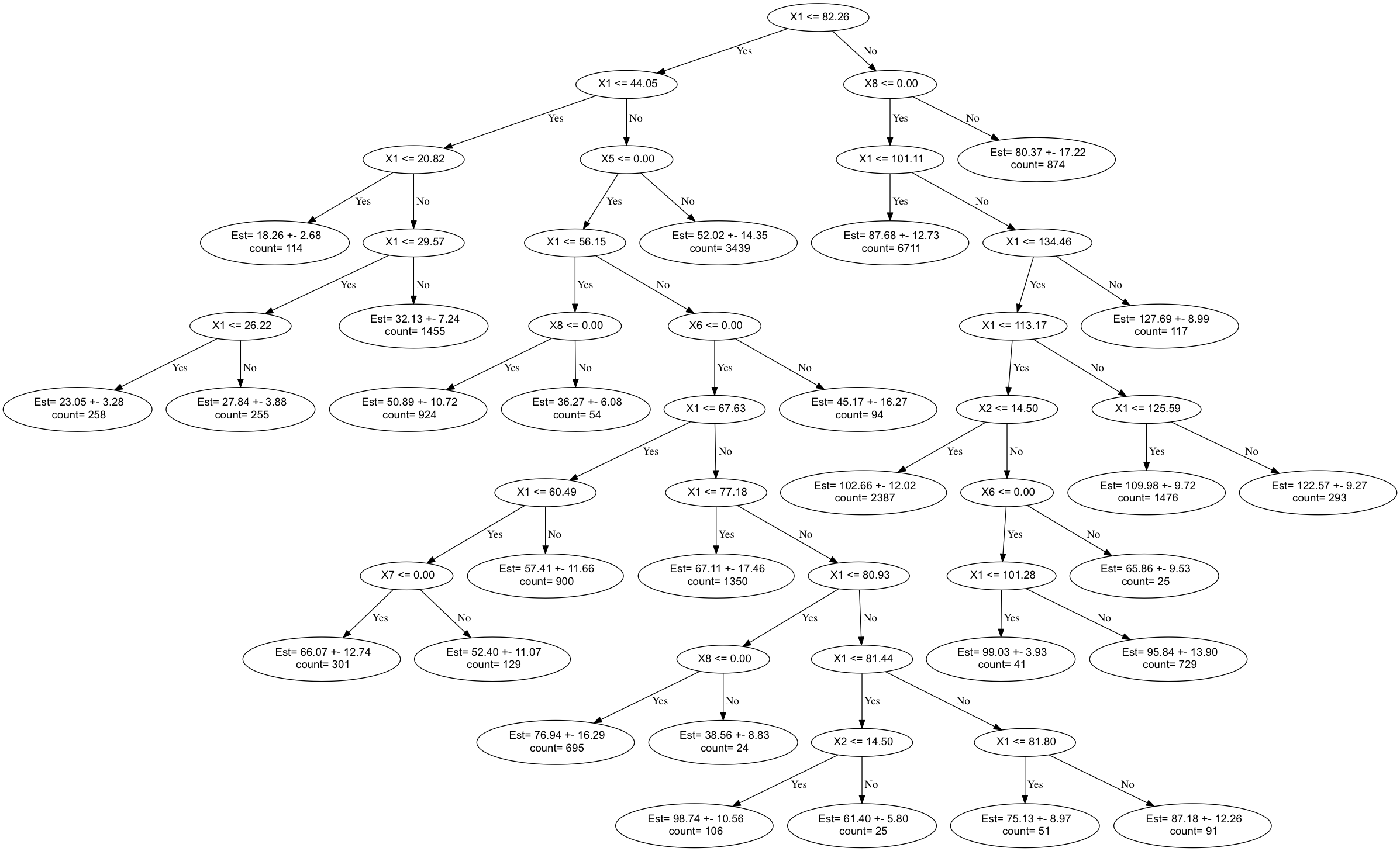}
	\caption{Cluster 2 found by the BET model applied on the cystic fibrosis data}
	\end{figure}

\end{appendix}

\end{document}